\documentclass[10pt,twocolumn,letterpaper]{article}

\usepackage{wacv}      

\usepackage{graphicx}
\usepackage{amsmath}
\usepackage{amssymb}
\usepackage{booktabs}
\usepackage{svg}

\usepackage[colorinlistoftodos,prependcaption,textsize=tiny]{todonotes}

\usepackage[pagebackref,breaklinks,colorlinks]{hyperref}

\usepackage[capitalize]{cleveref}
\crefname{section}{Sec.}{Secs.}
\Crefname{section}{Section}{Sections}
\Crefname{table}{Table}{Tables}
\crefname{table}{Tab.}{Tabs.}

\hypersetup{
    colorlinks=true,
    linkcolor=blue,
    filecolor=magenta,      
    urlcolor=cyan,
    pdftitle={Overleaf Example},
    pdfpagemode=FullScreen,
    }

\def\wacvPaperID{1656} 
\def\confName{WACV}
\def\confYear{2024}

\title{
    OPTIMUS: Observing Persistent Transformations in Multi-temporal Unlabeled Satellite-data
}
\author{
    Raymond Yu$^{*1}$, Paul Han$^{*1}$, Josh Myers-Dean$^2$, Piper Wolters$^2$, and Favyen Bastani$^2$\\
    \\
    $^1$University of Washington \\
    $^2$Allen Institute for AI\\
    \texttt{\small \{ryu5, paulh27\}@cs.washington.edu, \{joshm, piperw, favyenb\}@allenai.org}
}
\date{}

\begin{document}

\maketitle
\renewcommand{\thefootnote}{\fnsymbol{footnote}} 
\footnotetext[1]{Equal contribution.}
\renewcommand{\thefootnote}{\arabic{footnote}} 


\begin{figure*}
    \centering
    \includegraphics[width=\textwidth]{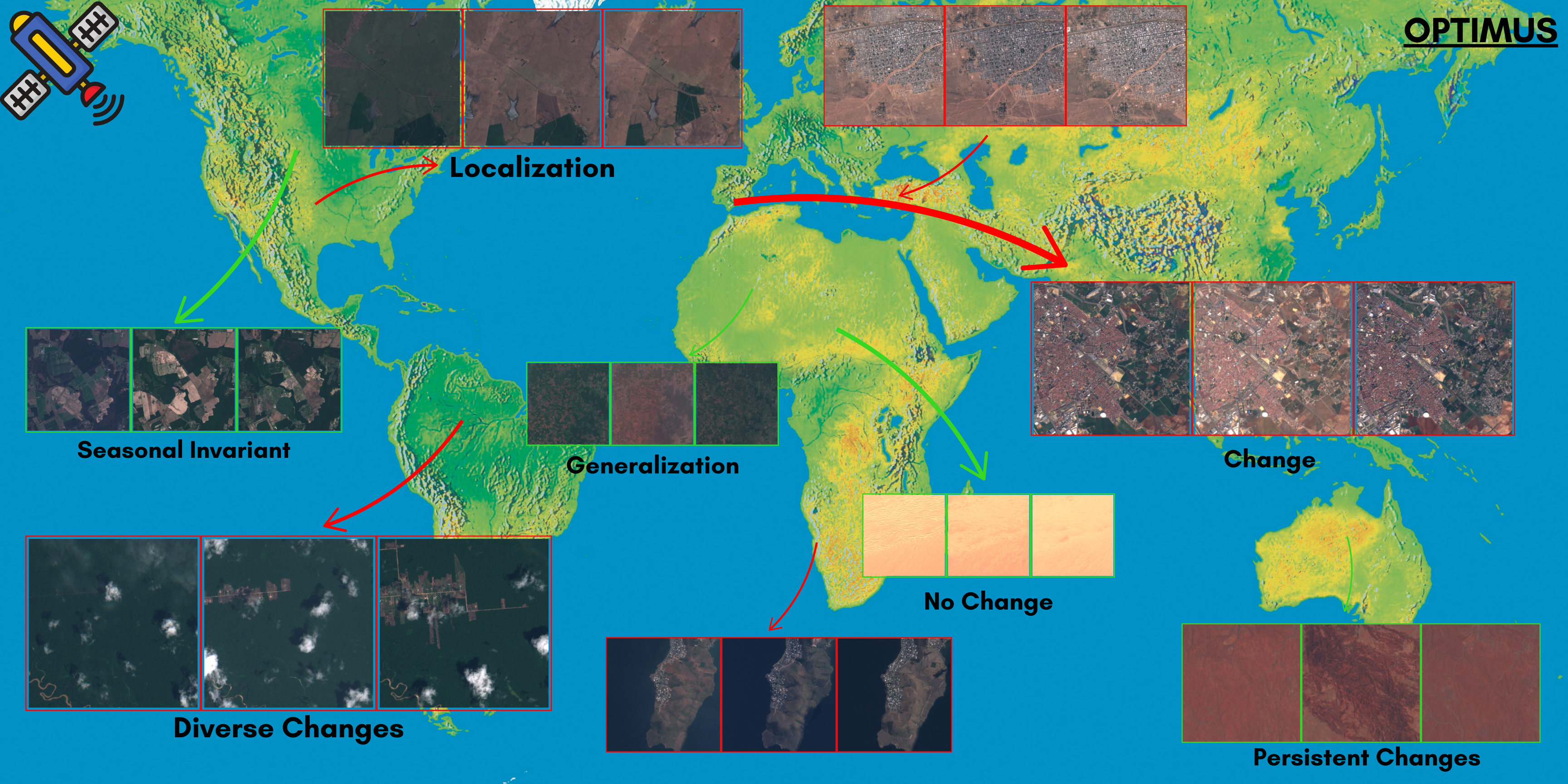}
    \caption{Overview of binary outputs (red for change, green for no change) generated by OPTIMUS across global regions. The figure showcases key capabilities: localization of changes into 128x128 images (top left) for precise spatial detection, identification of diverse changes such as deforestation in the Amazon Rainforest (red, label = 1) and shifting sand dunes in Chad (green, label = 0), and generalization across varied contexts (middle), including urban expansion in a city (red, label = 1) and lake recession due to drought (red, label = 1). OPTIMUS also demonstrates robustness to seasonal invariance, correctly identifying stable conditions such as vegetation changes in agricultural regions (green, label = 0).}
    \label{fig:examples}
\end{figure*}



\begin{abstract}
   In the face of pressing environmental issues in the 21st century, monitoring surface changes on Earth is more important than ever. Large-scale remote sensing, such as satellite imagery, is an important tool for this task. However, using supervised methods to detect changes is difficult because of the lack of satellite data annotated with change labels, especially for rare categories of change. Annotation proves challenging due to the sparse occurrence of changes in satellite images. Even within a vast collection of images, only a small fraction may exhibit persistent changes of interest. To address this challenge, we introduce OPTIMUS, a self-supervised learning method based on an intuitive principle: if a model can recover information about the relative order of images in the time series, then that implies that there are long-lasting changes in the images. OPTIMUS demonstrates this principle by using change point detection methods on model outputs in a time series. We demonstrate that OPTIMUS can directly detect interesting changes in satellite images, achieving an improvement in AUROC score from 56.3\% to 87.6\% at distinguishing changed time series from unchanged ones compared to baselines. Our code and dataset are available at \hyperlink{https://huggingface.co/datasets/optimus-change/optimus-dataset/}{https://huggingface.co/datasets/optimus-change/optimus-dataset/}.
\end{abstract}

\section{Introduction}

\label{sec:intro}
Globally detecting long-lasting changes to the Earth's surface is critical for informing decisions around tackling looming environmental, climate, and conservation challenges~\cite{HUSSAIN201391,8518015,rs11161854,gupta2019xbddatasetassessingbuilding}. Monitoring deforestation helps to understand where and why forest loss is happening; tracking urban expansion helps to quantify the environmental effects of urban sprawl; and identifying areas impacted by natural disasters like wildfires and earthquakes helps to target disaster relief efforts. Satellite imagery offers frequent views of any location on Earth; for example, the European Space Agency's Sentinel-2 mission provides images of land and coastal locations across the globe every five days.

Numerous computer vision datasets~\cite{daudt2018urban,shi2021deeply,chen2020spatial} and methods have been proposed for detecting changes in the tens of millions of square kilometers of the Earth that are imaged daily. However, the vast majority of these datasets focus narrowly on urban changes like new buildings and roads. This focus arises because annotating other types of long-lasting changes is expensive due to their rarity: most terrestrial locales exhibit minimal change over time, so even within a sizable dataset of satellite image time series, only a handful of time series may show certain changes like forest loss or the impact of wildfires. Furthermore, annotation demands specialized knowledge, as interpreting satellite images (especially outside urban areas) is not always straightforward. Additionally, the sheer number of potential categories of change makes it infeasible to develop a truly comprehensive dataset.

Because of these challenges—high annotation costs, the rarity of substantial changes, and the specialized knowledge required—many researchers have resorted to unsupervised methods for detecting changes in satellite image time series. A few unsupervised methods have been proposed, such as CaCo~\cite{caco-23}, which learns representations that diverge when a location undergoes change. However, these methods struggle to distinguish seasonal changes, such as deciduous trees changing color or crop fields being harvested, from long-lasting changes that permanently alter the Earth's surface.

To address these challenges, we propose OPTIMUS (Observing Persistent Transformations in Multi-temporal Unlabeled Satellite-data), a self-supervised learning method for classifying the presence of long-lasting change in satellite image time series. We focus on persistent changes, i.e., changes that have a visible impact on a location lasting longer than one year. OPTIMUS is based on the intuition that if a model can determine the correct long-term ordering of images in a time series, even when the sequence is hidden, it indicates the presence of significant, persistent changes in the images. Otherwise, if the location remained constant, the long-term ordering should be indecipherable. OPTIMUS only requires a collection of unlabeled satellite image time series for training.\\ 

Given a collection of image time series, OPTIMUS produces the subset of time series that underwent some long-lasting change. This enables users interested in analyzing changes in certain geographies to hone in on the small subset of patches within that geography that changed. In addition to classifying the presence of long-lasting changes, we show that simple extensions enable OPTIMUS to localize the changes both in space and in time within the image time series. Although OPTIMUS does not categorize the changes that it detects, category-specific annotation is substantially cheaper over the filtered subset, making it feasible to develop models that identify rare types of changes. Thus, OPTIMUS can be used to equip decision-makers with reliable data on the locations of changes like wildfires, flooding, deforestation, and so on, to help address and mitigate the impacts of global environmental challenges.

We developed a dataset consisting of one million image time series for training, along with a set of time series that we annotated with binary classification labels (has change or no change).
Our zero-shot results on the test set show that OPTIMUS significantly outperforms other unsupervised methods, improving the AUROC score at distinguishing changed time series from 56.3\% to 87.6\%. OPTIMUS helps users hone in on locations with persistent changes, allowing for a smaller, more manageable subset of images that can then be manually categorized by experts. This allows OPTIMUS to be a solution for wide-reaching applications in environmental monitoring, urban planning, disaster response, and conservation efforts.

\begin{figure*}[t]
    \centering
    \includegraphics[width=\linewidth]{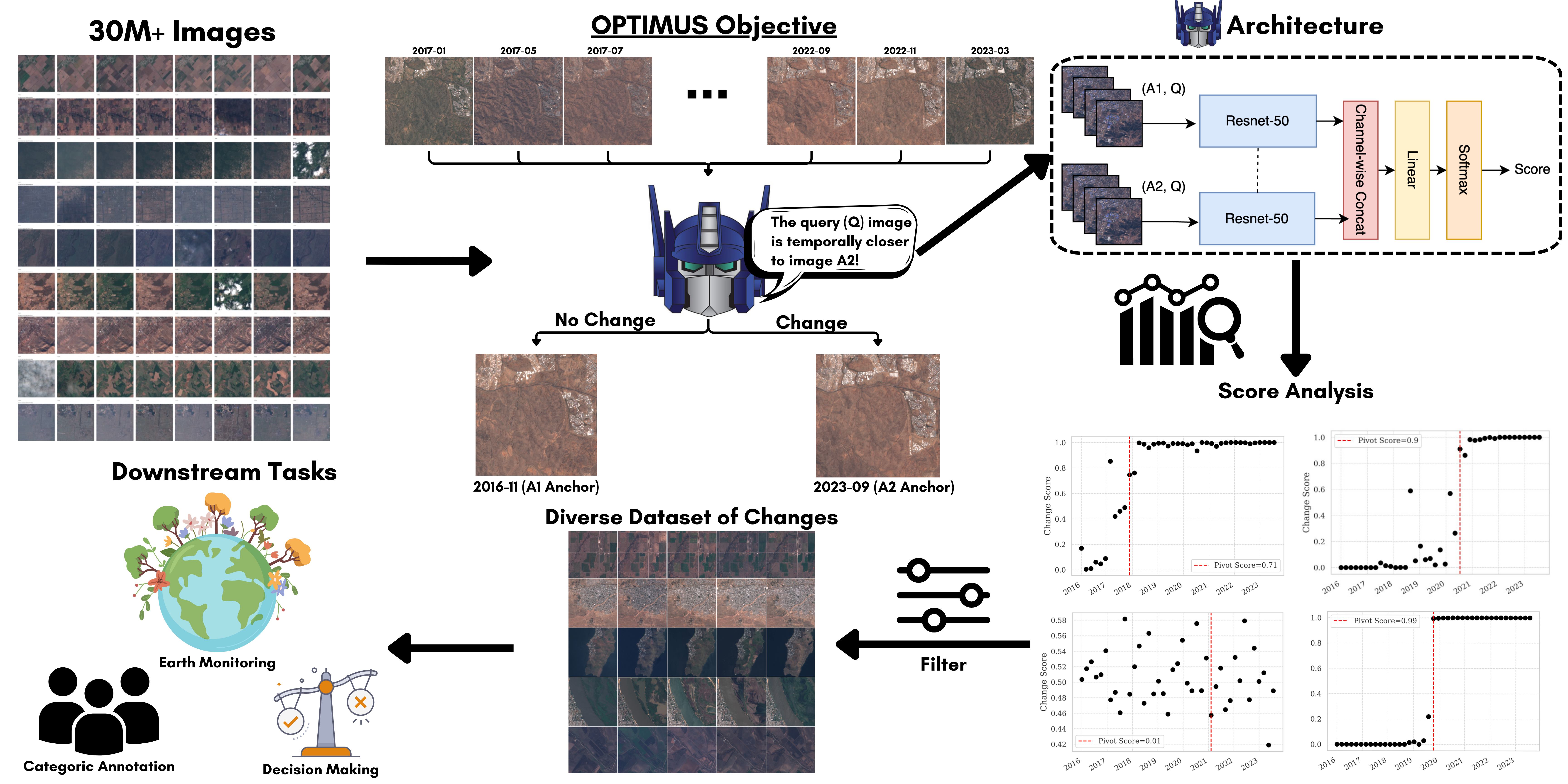}
    \caption{Overview of the OPTIMUS framework for detecting persistent changes in multi-temporal satellite imagery. The framework processes over 30 million satellite images, leveraging a Siamese network architecture to classify whether a query image (Q) is temporally closer to one of two anchor images (A1 or A2). This self-supervised approach generates a diverse dataset of annotated changes, distinguishing "Change" and "No Change" events while filtering out seasonal variations. Downstream tasks include environmental monitoring and decision-making, enabling actionable insights from large-scale satellite data.}
    \label{fig:tcp_model_summary}
\end{figure*}

\section{Related Works}

\subsection{Existing Datasets}
Several change detection datasets have been released, combining bitemporal satellite images or image time series with patchwise or pixelwise change labels.
OSCD~\cite{daudt2018urban}, Cropland-CD~\cite{liu2022cnn}, and SYSU-CD~\cite{shi2021deeply} provide binary pixelwise change masks for bitemporal images, with most labels corresponding to new buildings, new roads, and construction work.
Some datasets focus on individual categories.
EGY-BCD~\cite{holail2023afde}, SI-BU~\cite{liao2023bce}, and S2Looking~\cite{shen2021s2looking} have multiclass pixelwise change labels for constructed and demolished buildings.
Hi-UCD~\cite{tian2020hi} consists of labels for nine categories of change in images of Tallinn, including newly constructed buildings, greenhouses, and roads.

While these datasets are invaluable for change detection, they often focus on urban areas due to the relative ease of detecting changes in these regions. Acquiring labeled data for more complex environmental changes, like forest loss, wildfires, or desertification, is both costly and labor-intensive. These changes are subtle, long-term transformations that require specialized knowledge to detect and annotate accurately, unlike the more visually distinct changes in urban areas.

Some datasets provide multi-temporal data for land cover types.
DynamicEarthNet~\cite{toker2022dynamicearthnetdailymultispectralsatellite} provides monthly pixelwise labels for seven land use and land cover (LULC) classes, focusing on environmental changes like vegetation shifts and urbanization. DFC2021 Track MSD~\cite{9690575} targets land cover transitions using multisensor data fusion. However, these datasets are often limited due to the cost and are primarily centered around urban areas, making scalability and broader applicability an issue.


\subsection{Supervised Methods for Change Detection}



Supervised methods for change detection have demonstrated significant success in identifying changes in satellite imagery, but they rely on large amounts of labeled data, which presents scalability challenges. For instance, ChangeNet~\cite{ji2024changenetmultitemporalasymmetricchange} leverages deep neural networks with pre-trained weights, achieving high accuracy by fine-tuning on specific tasks. Despite its strong performance, ChangeNet requires extensive annotated datasets to generalize effectively, which is a notable limitation for broader applications. Similarly, the framework proposed by Wu et al~\cite{Wu2022FullyCC} focuses on analyzing image pairs, restricting its utility in detecting long-term changes across extended time series.

Additionally, many supervised methods, including Zheng et al's.~\cite{zheng2023changeeverywheresingletemporalsupervised} framework, are targeted towards urban change detection where data is more abundant. This urban-centric focus further restricts their broader environmental applicability of the models.


While pre-trained models~\cite{bastani2023satlaspretrain} can reduce the amount of labeled data needed for downstream tasks like change detection, supervised methods still face significant challenges in terms of generalizability and scalability due to the lack of labeled data, especially for rare categories of change like desertification, shrub regrowth, and logging.

\subsection{Unsupervised Methods for Change Detection}
Given the scarcity of labeled satellite data, unsupervised methods have gained traction in recent years. These approaches aim to detect changes without the need for annotated datasets, making them more scalable and applicable to various scenarios.

SeCo~\cite{mañas2021seasonalcontrastunsupervisedpretraining} is an unsupervised method pre-training method for remote sensing data that leverages temporal and positional invariance to learn transferable representations. Similarly, SSL4EO~\cite{wang2023ssl4eos12largescalemultimodalmultitemporal} utilizes self-supervised learning techniques to learn representations of satellite images that can be used to detect changes. Both SeCo and SSL4EO have shown effectiveness in identifying changes across diverse environments; however, they often require fine-tuning to optimize performance for specific tasks and datasets.

On the other hand, methods like CaCo~\cite{caco-23} do not require fine-tuning. CaCo uses contrastive learning to differentiate between unchanged and changed areas in satellite imagery by obtaining changes using features extracted from the images. Specifically, CaCo addresses the issue of varying seasonal and location-specific changes by normalizing the distance between feature representations of images. It calculates a ratio of distances between long-term images and short-term seasonal images to normalize scaling differences across different locations. This ratio helps distinguish actual changes from seasonal variations. Despite this sophisticated approach, CaCo struggles with effectively separating long-lasting changes from environments with pronounced seasonal changes due to its reliance on feature distance ratios, which may not always capture the true magnitude of persistent changes. 

Recent work by Mall et al.~\cite{change-events-22} introduces an event-driven change detection approach, which focuses on detecting meaningful change events from spatio-temporal satellite imagery in Cairo and California. Their method addresses the challenge of discovering significant changes from vast amounts of time-series data, similar to our objective of capturing persistent transformations. By focusing on events rather than pixelwise differences, this approach aligns with the need for methods that generalize across broader contexts.

Our approach differs from existing methods ~\cite{zheng2023changeeverywheresingletemporalsupervised} ~\cite{5169964} ~\cite{9555824} ~\cite{ZHANG2020183} by leveraging the full time series of satellite imagery in a fully unsupervised manner. This allows our model to better identify persistent changes, as it can learn from the temporal context provided by the entire series. By focusing on persistent changes, our methods aim to filter out large datasets, enabling experts to concentrate on images with significant and lasting alterations. This approach addresses some of the limitations of both supervised and unsupervised techniques, providing a novel method for identifying and monitoring significant environmental changes over time.



\section{Dataset}
Most existing change detection datasets, such as OSCD and ChangeNet, focus predominantly on urban areas. This urban-centric sampling is adopted to achieve a higher rate of detectable changes, making annotation more feasible. However, this approach limits the variety of changes captured, excluding many non-urban changes such as wildfires, desertification, and agricultural expansion.

To address these limitations, we compile a new dataset from publicly accessible sources, Sentinel-2 and NAIP, with the goal of detecting arbitrary persistent changes across diverse geographical locations. Unlike existing datasets, our approach involves randomly sampling image patches from across the entire globe, including non-urban areas, to ensure a broader spectrum of changes is captured. This strategy enables us to annotate a more diverse set of changes, avoiding the bias toward urban areas.

To make this global sampling feasible, we focus on classifying time series of image patches for change rather than segmenting individual pixels. This is particularly advantageous for capturing non-urban changes, where drawing precise segmentation labels can be challenging. For instance, changes such as wildfires and desertification can have diffuse boundaries that are difficult to delineate precisely. 

For model training, we retain only the RGB channels (B2, B3, B4) and applied a cloud cover filter to exclude images with greater than 20\% cloud cover. Each image in our dataset measures 512x512 pixels at a 10-meter/pixel resolution, allowing our models to detect fine-grained changes that may occupy only a few pixels. The dataset comprises one million time series, each representing the spatio-temporal context of a geographic location. Each time series includes 30-48 satellite images captured between January 2016 and December 2023, with a minimum interval of 2 months between any two images. This extensive temporal span of 8 years ensures the dataset captures a wide range of long-term changes, including deforestation and desertification, providing a robust basis for training models to detect persistent changes.

\subsection{Evaluation Set} To evaluate our change detection models, we constructed an evaluation set of 300 satellite time series, each consisting of 512x512 images, randomly sampled from our dataset. A random subset of 100 of these images were further divided into 1600 128x128 patches, with each patch receiving a binary label. Each time series in this set is assigned a binary label: positive labels indicate the presence of a persistent and non-cyclic change (i.e., not seasonal changes), while negative labels indicate no change. Approximately one-third of the time series are positively labeled.

Given the complexity and diversity of changes we aim to detect, we prioritized the quality of annotations over quantity. To ensure high-quality annotations, we provided annotators with detailed instructions on how to identify and label persistent changes, emphasizing consistency across different types of changes, especially those with ambiguous boundaries, such as desertification or gradual urban expansion. Annotators were instructed to focus on clear indicators of non-cyclic change, disregarding seasonal variations or temporary phenomena.

Appendix~\ref{appendix:annotation_examples} provides qualitative examples of the annotated changes, illustrating the types of changes our models are designed to detect. These examples are accompanied by the specific instructions given to annotators, offering insight into the criteria used for labeling. 

The evaluation task involves the models generating a change score for each time series in the evaluation set and computing the task accuracy. This evaluation closely mirrors practical downstream applications, where the objective is to identify locations with persistent changes over many years. Although using a binary label is a relatively simple approach—lacking the granularity of change categories or segmentation masks—it serves as an effective filtering mechanism and remains effective for types of changes like long-term droughts that have nebulous boundaries. This simplicity makes it cost-efficient to extract a diverse set of changes in future applications.

We will release the unsupervised training data, labeled test set, and code (which is also included in the supplementary material) under an open license.
\label{eval_set_annotation}


\section{Methods}
OPTIMUS determines whether remote sensing satellite time series contain non-seasonal changes by hiding the ordering of the images in the series, and then training a model to attempt to recover the long-term ordering. If there are only seasonal changes but no persistent changes, then the long-term ordering should be indecipherable from the images alone. An example is lake levels changing due to seasonal precipitation. As this change is cyclic, while the images can be grouped into seasons, the ordering of the images across years cannot be distinguished. On the other hand, if a location exhibits persistent changes, it should be possible to distinguish all of the images captured before a change from those captured after a change. Road construction is such an example, with distinct stages like laying the subgrade, base, and asphalt.

In this section, we first describe a basic implementation of this intuition. We then identify flaws in the basic implementation, and detail how OPTIMUS addresses those flaws.

In the basic implementation, given a time series $\langle I_1, I_2, \ldots, I_n \rangle$, we train a binary classifier (denoted as $b$) to predict whether an arbitrary intermediate image $Q$ is closer in time to $I_1$ or $I_n$. The classifier inputs a tuple $(I_1, I_n, Q)$ of the images only, with the timestamp of $Q$ hidden, and outputs a confidence score that $Q$ is closer to $I_n$ than $I_1$. During training, examples are constructed by (1) sampling a time series, (2) picking an arbitrary $Q$ between $I_1$ and $I_n$, and (3) computing the label based on whether $Q$ is before or after $I_{n/2}$.

After training, to determine if a new time series contains change, we compute the confidence score from the model for every image in the time series, i.e. we compute a time series of scores $S = \{b(I_1, I_n, I_j) | j = 1, 2, \ldots, n\}$. An oracle classifier would output a step function where the score switches from 0 to 1 halfway through the time series. If the time series contains change, then an effective model should perform similarly to the oracle (Figure \ref{fig:tcp_model_graph}, top left). However, if the time series contains no persistent changes, then the scores should fluctuate arbitrarily, since the model does not have sufficient information (Figure \ref{fig:tcp_model_graph}, bottom left). Then, we apply measures on $S$ that broadly assess the degree to which it is monotonically increasing. 

However, there are several flaws with this basic implementation. First, a location may undergo a change at a single timestamp, e.g. trees are logged within the span of a month. Suppose this change occurs between $I_1$ and $I_2$. Then, during training, we would be training the model to predict that $\langle I_2, ..., I_{n/2-1} \rangle$ are closer in time to $I_1$, even though they are closer in appearance to $I_n$. Second, if $I_1$ or $I_n$ are low in quality, due to clouds, shadows, or imaging artifacts, then the quality of the scores in $S$ would be lowered. Below, we address these flaws, and also detail the measures that we use to capture a final change score based on the confidence scores in $S$, along with the model architecture.

\smallskip
\noindent
\textbf{Training Example Selection.}
\label{train_and_arch}
To train the classifier, we sample triplets $(A_1, A_2, Q)$ for each time series in the dataset, and train the model to predict whether $Q$ is closer to $A_1$ or $A_2$. Selecting $A_1 = I_1$, $A_2 = I_n$, and $Q$ as a random image from within the time series is the simplest approach, but has flaws as mentioned above.

The first concern is that a change may occur just after $A_1$ or just before $A_2$, causing a randomly chosen $Q$ to be temporally closer to one anchor but visually more similar to the other. To address this, during training, rather than selecting a query image $Q$ between $A_1$ and $A_2$, we select $Q$ to be before $A_1$ or after $A_2$ with equal probability; this guarantees that, with respect to persistent changes, $Q$ will always be visually more similar to the same anchor that it is temporally closer to.

Second, to address the concern of low quality images due to environmental or photometric distortions, rather than using a single image for the anchors $A_1$ and $A_2$, we actually provide the model with multiple consecutive images. Specifically, each anchor consists of $c = 3$ consecutive satellite images from the time series. Using multiple consecutive images for $A_1$ and $A_2$ reduces the likelihood that all images in a set will be affected by distortions, increasing the robustness of the model. However, there is a trade-off: if the anchor sets are too long, there is a risk that changes may occur within $A_1$ or $A_2$ themselves, potentially complicating the model's learning process. Through ablation studies, we determined that using three consecutive images strikes an optimal balance, providing sufficient robustness while minimizing the risk of internal changes within the anchor sets. More details on the input construction can be found in the Appendix \ref{appendix:ablations}.

\smallskip
\noindent
\textbf{Model Architecture.}
For each triplet $(A_1, A_2, Q)$, we construct two tensors that the model processes independently. The first tensor is formed by concatenating the query image $Q$ with each image in the anchor $A_1$, resulting in an $3 \times (c+1) \times 512 \times 512$ tensor. Similarly, the second tensor is created by concatenating $Q$ with the images from anchor $A_2$. These tensors are denoted as $(A_1, Q)$ and $(A_2, Q)$, respectively. They are then input to the model, which is tasked with predicting whether the query $Q$ is temporally closer to $A_1$ or $A_2$. The model is trained using binary cross-entropy loss, defined as:
\[ \omega =   b((A1, Q), (A2, Q))\]
\[L_\text{OPTIMUS} = -y \log (\omega) - (1-y) \log (1 - \omega)\]
where $\omega$ is the output of the binary classifier and $y \in \{0,1\}$ is the ground truth label indicating whether $Q$ is closer to $A_1$ ($y = 0$) or $A_2$ ($y = 1$).

The model architecture uses a Siamese Neural Network \cite{siamese2015} with a Resnet-50 backbone \cite{he2015deepresiduallearningimage} to map the tensors into a shared embedding space. The embeddings are concatenated, passed through a linear layer, and then processed by a softmax function to yield a score between 0 and 1. The model is trained using the AdamW \cite{loshchilov2019decoupledweightdecayregularization} optimizer with a learning rate of $3 \times 10^{-4}$, a batch size of 5, and for 5 epochs. The dataset is partitioned into training and validation sets with an 80/20 split. The training objective and model architecture are depicted in Figure \ref{fig:tcp_model_summary}.

\smallskip
\noindent
\textbf{Change Score Measures.}
Given a time series $\langle I_1, I_2, ..., I_n \rangle$, suppose we calculated the series of scores $S = \{b(I_1, I_n, I_j) | j = 1, 2, \ldots, n\}$. We use two measures to quantify persistent changes in the sequence. The first is the Spearman rank correlation coefficient:
$$\rho = 1 - \frac{6 \sum_{i=1}^{|S|} (\text{rank}(s_i) - i)^2}{|S|(|S|^2 - 1)}$$
This metric assesses the monotonicity of $S$ by measuring how well the ranks of the  scores correspond to their positions in the sequence. The second measure is the pivot score $P$:
$$ P = \max_{i=1}^{n} \left| \frac{\sum_{j=1}^{i} s_j}{i} - \frac{\sum_{j=i+1}^{n} s_j}{n-i} \right| $$
This score identifies the index $i$ in $S$ that maximizes the absolute difference between the average values of the segments before and after $i$. A high pivot score indicates a significant and persistent change in the sequence, typically aligning with abrupt transitions. For a visualization of the scoring process, see Figure \ref{fig:tcp_model_graph}.

In practice, as shown by an ablation study in  Appendix \ref{appendix:ablations}, the pivot score outperformed the Spearman coefficient, so all subsequent results and experiments report scores based on the pivot score.

\subsection{Spatial Localization of Changes}
\label{iterative_training_prod}
OPTIMUS operates effectively as a general framework for large 512$\times$512 images, corresponding to a 5km$\times$5km region at 10m resolution. However, this extensive spatial context complicates the precise localization of regions or pixels where changes occur. For instance, urbanization might be confined to a small section of the image due to the construction of a few buildings. Running OPTIMUS on smaller patches to localize changes in a time series significantly degrades performance because the model was trained on large spatial contexts and cannot generalize well to smaller patches.

A straightforward solution is to train the model on 128x128 patches to adapt to the lower spatial context. However, this approach introduces the challenge of sparse changes when subdividing time series into smaller patches. Most regions in an image do not exhibit change, resulting in a low signal-to-noise ratio during model training.

To address this issue, we propose an iterative approach to extend OPTIMUS for spatial localization. Initially, we train OPTIMUS on the entire dataset. We then apply the trained model to filter the top 50\% of time series that show the most persistent changes, thereby increasing the density of changes in the dataset and enhancing the model's ability to learn from significant changes. Subsequently, we retrain OPTIMUS on this filtered dataset using 128$\times$128 patches.




\section{Experiments}
\begin{figure*}
    \centering
    \includegraphics[width=1\linewidth]{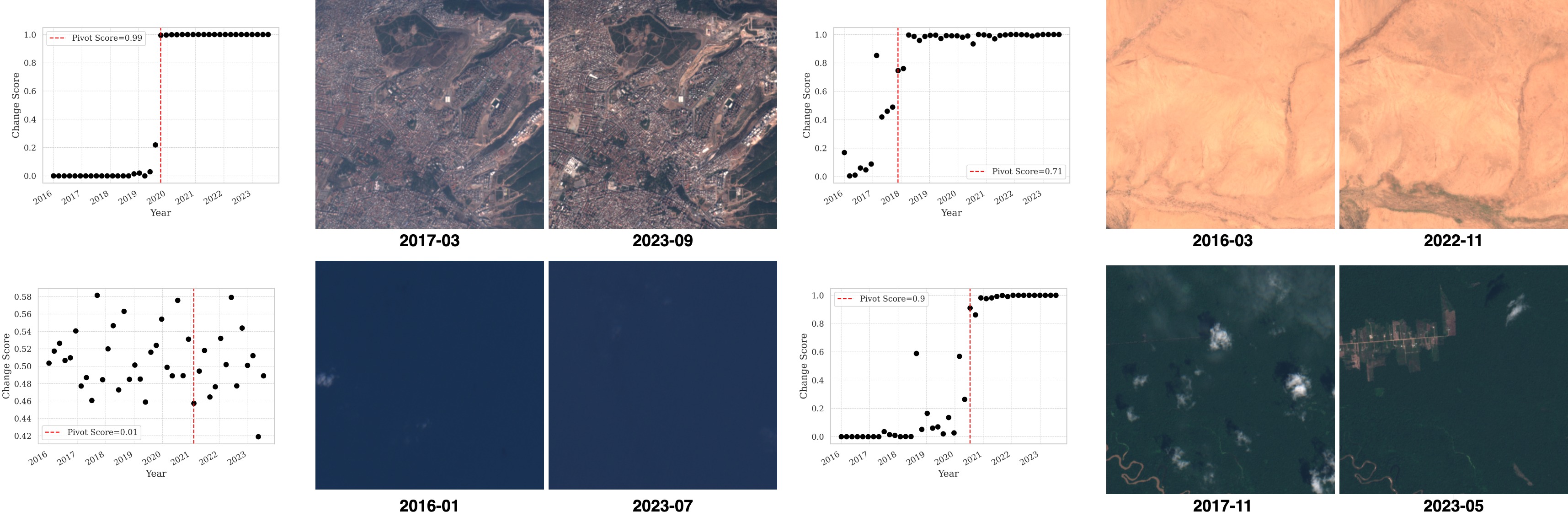}

    \caption{Analysis of pivot scores and corresponding satellite images for different environmental changes. The top left shows a high pivot score indicating significant urban expansion with a marked spike in change, suggesting a sudden development. The bottom left illustrates ocean images associated with an extremely low pivot score, indicating minimal change. The top right depicts a desert region experiencing reforestation, with shrubs growing back. The bottom right shows a region undergoing deforestation.}
    \label{fig:tcp_model_graph}
\end{figure*}

\subsection{Change Detection}
We first evaluate OPTIMUS against baselines on distinguishing $512\times512$ time series exhibiting persistent changes from those with no changes in our evaluation set.\\\\
\smallskip
\noindent
\textbf{Baselines.} We compare OPTIMUS to three baselines: CaCo, SeCo, and OSCD pre-training.

CaCo is a contrastive method specifically designed to detect long-term changes while being insensitive to seasonal variations. The authors apply CaCo for classifying changes by computing the distance between feature representations. To classify a time series using CaCo, we randomly sampled three image pairs from the first and last years, then computed the average distance between their representations. A larger distance indicated a greater change score, relative to the other time series. Following the authors' recommendations, we also normalized the distances between feature vectors representing long-term and short-term images, as the feature distances can scale differently depending on the type of location.

CaCo~\cite{caco-23} is the only recent method geared for change classification, since most methods focus on segmentation. Thus, we make adaptations to compare the other two baselines. 

Unlike CaCo, SeCo~\cite{mañas2021seasonalcontrastunsupervisedpretraining} is a contrastive model intended for seasonally variant and invariant downstream applications. As such, it produces embeddings with three independent subspaces, including one variant and another invariant to seasonal changes. Since our task focuses on detecting non-seasonal changes, we explicitly remove the seasonally variant subspace and only use the other two subspaces to quantify changes. Following this, however, the evaluation method for SeCo is kept the same as CaCo.

The third baseline consists of pre-training on OSCD ~\cite{daudt2018urban}, a dataset consisting of bi-temporal images paired with change masks. After pre-training, we compare feature representations to classify change. Since CaCo reports the highest performance on OSCD, we use it for the model, fine-tuning it on OSCD following the authors' training procedure. To evaluate on our dataset, change scores for each time series are calculated as before, using distances between representation vectors.

\begin{table}[ht]
\centering
\caption{Comparative Evaluation versus Baseline Methods}
{\footnotesize
\begin{tabular}{@{}l@{\hspace{6pt}}c@{\hspace{6pt}}c@{\hspace{6pt}}c@{\hspace{6pt}}c@{}}
\toprule
\textbf{Method}  & \textbf{F1 Score} & \textbf{AUROC} \\
\midrule
SeCo &  0.540  & 0.491 \\
CaCo &  0.551 & 0.494 \\
CaCo (OSCD) & 0.585 &  0.563  \\
OPTIMUS & \textbf{0.760} &\textbf{0.876} \\
\bottomrule
\end{tabular}
}
\label{tab:bcc_evalulation}
\end{table}

\begin{figure*}
    \centering
    \includegraphics[width=0.90\linewidth]{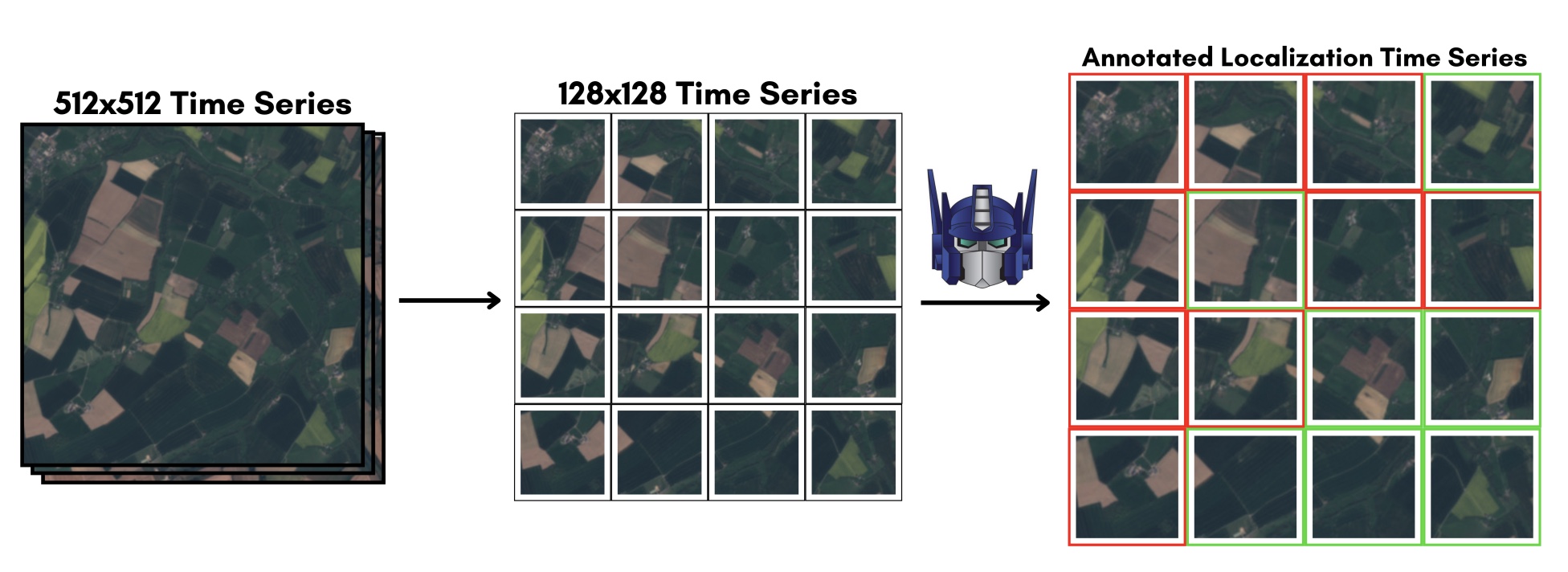}
    \caption{ Illustration of the proposed pipeline for fine-grained change detection. Satellite image time series are divided into 128x128 patches to isolate localized changes, enabling the OPTIMUS model to classify each patch as exhibiting meaningful changes (e.g., urban expansion) or no significant change (e.g., seasonal variations). This approach addresses challenges in detecting changes across diverse components within larger images by focusing on spatially localized transformations.}
    \label{fig:tcp_model_summary}
\end{figure*}

\smallskip
\noindent
\textbf{Metrics.}
For each baseline method, a change score was computed for each series in the evaluation set, and a threshold was applied to distinguish predicted changes from unchanged series. Performance was then measured in terms of accuracy against ground truth labels. Given the varying ranges of score values produced by different methods, the primary performance metric used is the non-thresholded Area Under the Receiver Operating Characteristic Curve (AUROC). In addition, we report the maximum F1 score across all thresholds.

\smallskip
\noindent
\textbf{Quantitative Results.}
Table \ref{tab:bcc_evalulation} shows the performance metrics for all evaluated methods on the evaluation set. OPTIMUS achieves a substantial performance advantage over all other methods.  We attribute this to OPTIMUS's approach of explicit classification of images based on temporal locality, which directly encodes changes into the feature representations. However, contrastive methods such as SeCo and CaCo only indirectly capture changes through differences in feature representations, resulting in less reliable outcomes. Lastly, supervised pretraining on OSCD increases the performance for CaCo, but it still falls far short of OPTIMUS. This gain can be attributed to CaCo's direct training for urban change detection, which helps detect urban images in our evaluation set. However, the overall effectiveness is limited due to the broader range of changes in our evaluation dataset beyond urban environments.

\subsection{Localization}
In Table \ref{tab:localization}, we present the results of the localized evaluation for OPTIMUS. Various versions of OPTIMUS are compared to demonstrate the necessity of the iterative training procedure outlined in Section \ref{iterative_training_prod}.  The original OPTIMUS model, trained on 512$\times$512 images, performs poorly with an AUROC score when applied to smaller patches. Training OPTIMUS directly on 128$\times$128 images without the iterative procedure results in an improvement in AUROC. However, the best performance is achieved with OPTIMUS trained using the iterative method, which increases AUROC significantly. This improvement is expected, given that the iterative OPTIMUS trains on a dataset with a much higher density of changes. The fairly high AUROC for the iterative OPTIMUS method indicates its effectiveness in localizing regions of interest for potential persistent changes. While the overall performance decreases compared to 512$\times$512 images, this is anticipated due to the reduced spatial context.



\begin{table}[ht]
\centering
\caption{Comparative Localized Evaluation}
{\footnotesize
\begin{tabular}{@{}l@{\hspace{6pt}}c@{\hspace{6pt}}c@{\hspace{6pt}}c@{\hspace{6pt}}c@{}}
\toprule
\textbf{Method} & \textbf{F1 Score} & \textbf{AUROC} \\
\midrule
OPTIMUS (512x512)  & 0.587 & - \\
OPTIMUS (128x128)  & 0.615 & 0.8714\\
OPTIMUS (512x512 $\rightarrow$ 128x128 iterative)  & \textbf{0.658} & \textbf{ 0.9415} \\
\bottomrule

\end{tabular}
}
\label{tab:localization}
\end{table}

\label{localization_section}

\subsection{Qualitative Examples}
Figure \ref{fig:tcp_model_graph} provides examples of change score series generated by OPTIMUS on the evaluation dataset, along with the outputs of the pivot scores. It also shows the exact time at which the pivot score detected the greatest degree of change. The results illustrate that the OPTIMUS accurately measures the degree of change and can often pinpoint precisely the location of the greatest change, where the partitioned sets of images before and after the pivot exhibit the greatest disparity. Qualitative examples are located in Appendix \ref{appendix:qual}.  

\section{Conclusion}
Change detection in remote sensing is essential for applications such as monitoring deforestation, tracking urban expansion, and assessing the impact of natural disasters. However, supervised methods struggle with the scarcity and high cost of obtaining labeled data, particularly for non-urban areas. In this paper, we introduced OPTIMUS, an unsupervised approach that identifies persistent, long-term changes in satellite image time series by learning temporal patterns. OPTIMUS improves the AUROC score from 56.3\% to 87.6\%, significantly outperforming previous methods in distinguishing significant changes from seasonal variations.
{\small
\bibliographystyle{ieee_fullname}
\bibliography{egbib}
}

\clearpage

\appendix
\section{Annotation Guidelines}
\label{appendix:annotation_examples}
For the manual annotation process, we provided annotators with a detailed set of guidelines to ensure consistency in labeling persistent changes. Annotators were instructed to focus on long-lasting changes that persisted for over a year, disregarding short-term or seasonal variations. They were given examples of desertification, urban expansion, and forest loss, and asked to ignore temporary changes such as crop rotations or seasonal foliage variations. If either annotator encountered uncertainties during the labeling process, we would review the time series, discuss our observations, and reach a consensus. We found that inter-annotator agreement was high, particularly in cases of clear, persistent changes. Furthermore, as the dataset is fairly small, we randomly sampled and double-checked each of the annotated time series.

Initially, we manually annotated 300 images of size 512x512 pixels with binary labels indicating whether persistent changes were present (1) or not (0). For images labeled as 0 (no change), we made the assumption that changes were uniformly absent throughout the image. As a result, these images were split into 16 smaller patches, each of size 128x128 pixels, and all patches were automatically labeled 0. For images labeled as 1 (indicating changes), all 16 patches were individually annotated to capture the finer details of the changes across smaller regions.

\begin{figure}[h]
  \centering
  \includegraphics[width=0.9\linewidth]{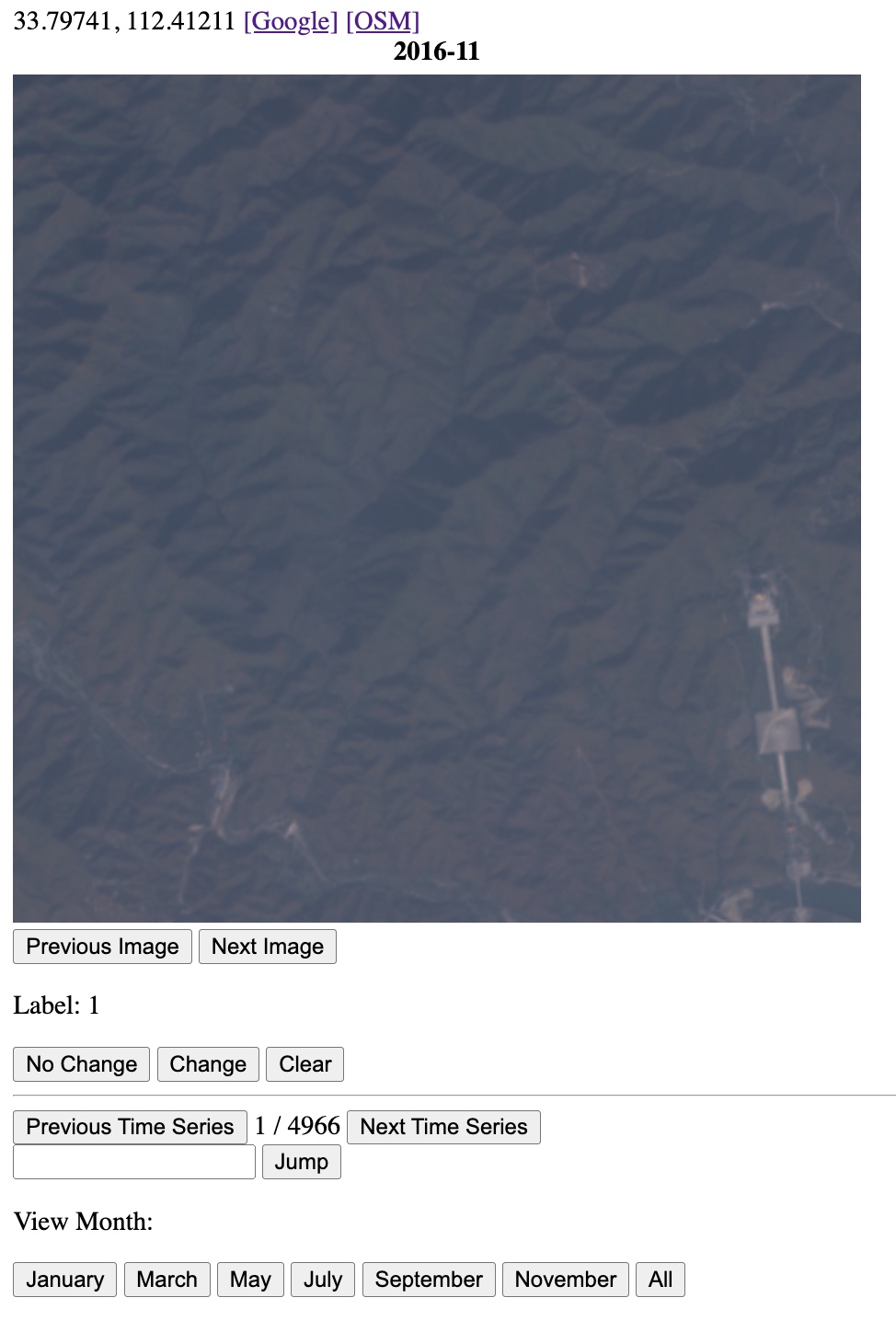}
  \caption{The annotation interface used during the manual labeling process. Annotators could navigate through time series images, view specific months, and classify changes using a set of keyboard shortcuts. Furthermore, they could adjust the number of images they wanted to view at a time (i.e., 3 at a time instead of 1).}
  \label{fig:annotation_interface}
\end{figure}

In addition to the manual annotation process, annotators had access to longitude and latitude information for each image, along with integrated Google Maps and OpenStreetMap views within the annotation interface (shown in Figure~\ref{fig:annotation_interface}). This integration allowed them to cross-reference geographical context, improving their ability to identify persistent changes. For example, if annotators were unsure about changes in an image, they could determine that the area was in Australia and recognize that fires between certain months may have affected the region. This spatial context greatly improved annotation accuracy for geographically complex or ambiguous cases.

\section{Qualitative Examples and Model Comparison}
\label{appendix:qual}
\begin{figure*}[h]
  \centering
  \includegraphics[width=\textwidth, height=0.95\textheight, keepaspectratio]{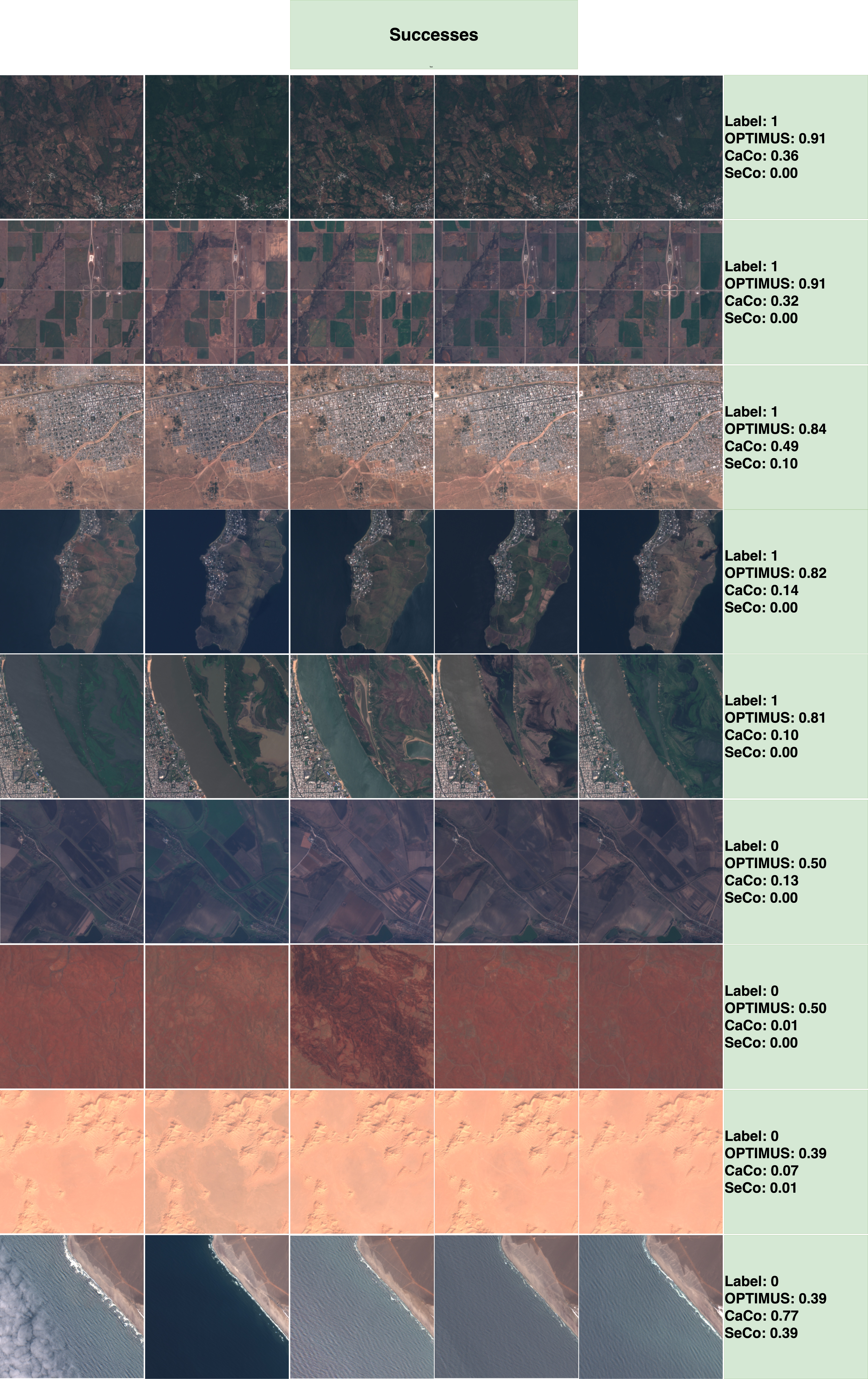}
  \caption{Examples of successful predictions by OPTIMUS, with performance scores from CaCo and SeCo models across various environments. The images are shown from left to right for the years 2016, 2018, 2020, 2021, and 2023, respectively, all captured in the month of November. Each row corresponds to a specific time series, with labels indicating whether a persistent, non-cyclic change is present (1) or absent (0).}
  \label{fig:positive}
\end{figure*}

The qualitative analysis of OPTIMUS, CaCo, and SeCo reveals important distinctions in how these models handle persistent and cyclic changes across diverse environments. OPTIMUS performs exceptionally well in detecting long-term, persistent changes by leveraging the temporal progression of images, allowing it to filter out short-term, cyclic variations like crop rotations or seasonal shifts in snow cover. As shown in Figure \ref{fig:positive}, OPTIMUS is effective in identifying clear, non-reversible transformations, making it suitable for both urban and natural environments where such variations dominate.

\begin{figure*}[h]
  \centering
  \includegraphics[width=\textwidth, height=0.95\textheight, keepaspectratio]{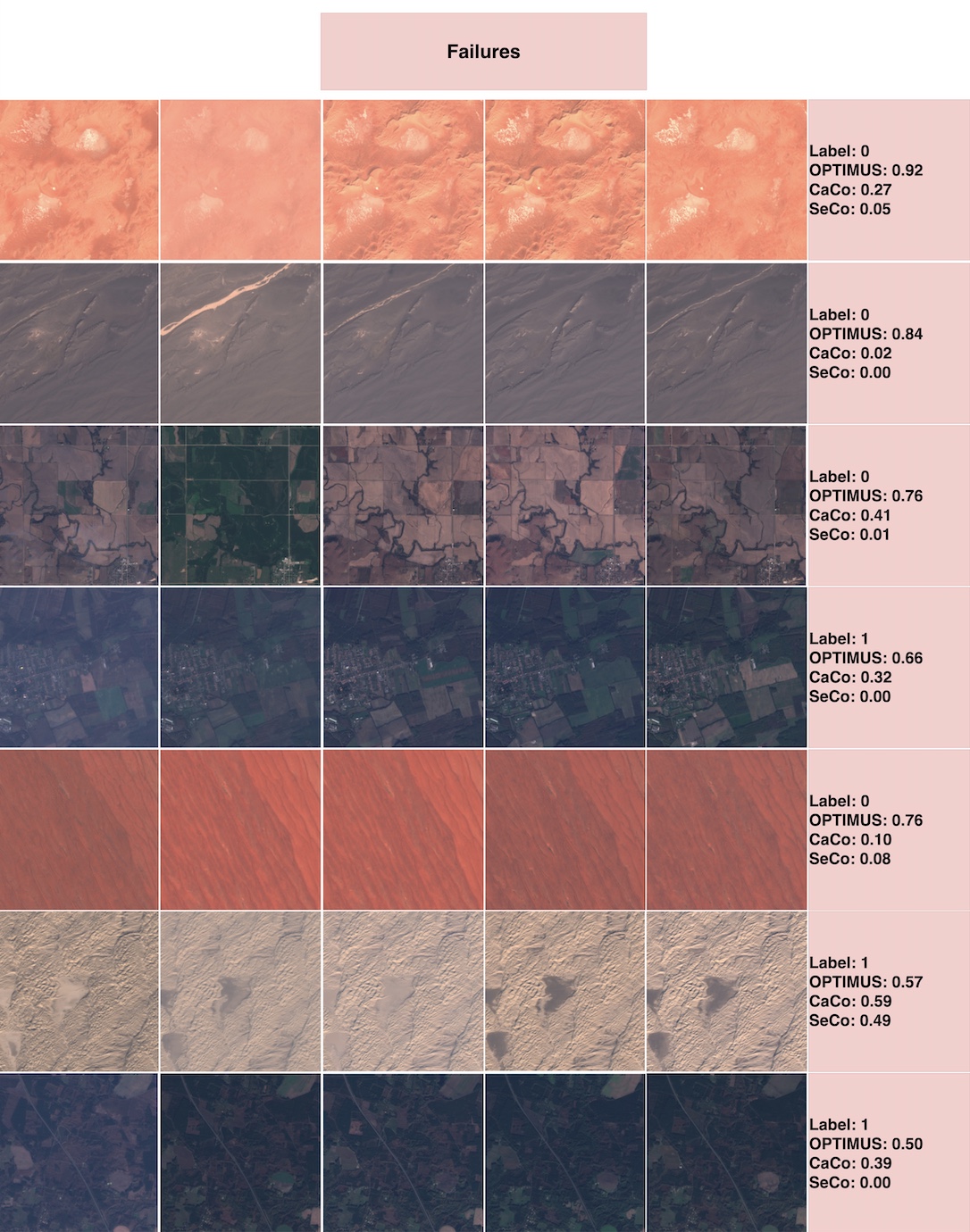}
  \caption{Most failures occur in complex environments with both seasonal and urban changes, or in situations where visual artifacts (e.g., shadows, lighting changes) mislead the models. Note that row 6 is incorrectly labeled (it should be 0); this was identified earlier, and we have since double-checked the evaluation dataset.}
  \label{fig:negative}
\end{figure*}

However, some failure cases highlight challenges for OPTIMUS in distinguishing between significant and subtle environmental shifts. In particular, regions like deserts or areas with strong lighting changes or surface texture shifts (e.g., sand dunes, shadows) can mislead the model, causing it to detect changes where none actually exist. These failure cases suggest that while temporal progression helps filter out short-term cyclic variations, certain natural phenomena—such as reflective surfaces or temporary shadows—can still trigger high change scores in OPTIMUS (see Figure \ref{fig:negative}). Notably, SeCo and CaCo exhibit similar patterns in these situations, as reflected in their segmentation masks.

CaCo also performs relatively well in rural environments due to its seasonally invariant representations. However, in our evaluation of both SeCo and CaCo, change scores are calculated by dividing the detected change by the total number of pixels in the image. This pixel-wise normalization makes them more effective in urban areas, where changes like new buildings or roads are concentrated and well-defined. In urban settings, larger and clearer change maps make the pixel-wise division less problematic. However, in rural environments where changes are smaller and more dispersed, this approach dilutes the change signal, making both SeCo and CaCo less sensitive to subtle transformations.

Many failure cases, as shown in Figure \ref{fig:negative}, occur in environments where multiple change signals overlap—such as urban expansion combined with cyclic agricultural changes. In these cases, the models must distinguish between permanent, meaningful changes and temporary cyclic phenomena. OPTIMUS generally performs better while SeCo has a tendency to output black change maps, indicating no detected change when the signal is weak or dispersed.

\section{Ablations}
\label{appendix:ablations}
For all ablations, we report the area under the ROC curve (AUROC), which is threshold-independent. Additionally, we include the optimal F1 scores across all thresholds.

\subsection{Backbones}
We evaluated various backbones to determine their impact on performance. The backbones tested include Resnet-50, Resnet-152, and Swinbase-v2~\cite{liu2022swintransformerv2scaling}. For these tests, we used Satlas ~\cite{bastani2023satlaspretrain} weights and a context size of three. Due to the availability of Satlas weights, we were limited to these three backbones.

\begin{table}[ht]
\centering
\caption{Testing different encoder backbones}
{\footnotesize
\begin{tabular}{@{}l@{\hspace{6pt}}c@{\hspace{6pt}}c@{\hspace{6pt}}c@{\hspace{6pt}}c@{}}
\toprule
\textbf{Backbone} & \textbf{F1 Score} & \textbf{AUROC} \\
\midrule
Resnet-50 & \textbf{0.760} &  \textbf{0.876} \\
Resnet-152 & 0.730 & 0.850 \\
SwinBase-v2  & 0.750 & 0.865 \\
\bottomrule
\end{tabular}
}
\label{tab:ablation_backbone}
\end{table} 

The AUROC performance was comparable across all tested backbones, suggesting that the core strength of our approach lies in the method of using change scores rather than the specific backbone used. Therefore, we opted for Resnet-50 due to its efficiency.

\subsection{Weight initializations}
We assess the effect of different weight initializations on the Resnet-50 backbone with a context size of three. We tested three initialization methods: Random, ImageNet ~\cite{5206848} ~\cite{bastani2023satlaspretrain}, and Satlas weights, as shown in Table \ref{tab:ablation_init}.

\begin{table}[ht]
\centering
\caption{Testing different initializations}
{\footnotesize
\begin{tabular}{@{}l@{\hspace{6pt}}c@{\hspace{6pt}}c@{\hspace{6pt}}c@{\hspace{6pt}}c@{}}
\toprule
\textbf{Initialization} &  \textbf{F1 Score} & \textbf{AUROC} \\
\midrule
Random &  0.723 & 0.851  \\
Imagenet  & 0.746 & 0.857\\
Satlas & \textbf{0.760} &  \textbf{0.876} \\
\bottomrule
\end{tabular}
}
\label{tab:ablation_init}
\end{table}

We acknowledge that this is not a comprehensive ablation analysis, as random and Imagenet weights were never ideal for the specific task. The primary aim was to demonstrate that using Satlas weights, which are pre-trained on satellite images, improves performance compared to Random and Imagenet weights.
\subsection{Context sizes}
For an ablation study, we evaluate the impact of context size on the performance of OPTIMUS, as described in Section \ref{train_and_arch}, by varying context size from one to five. For each configuration, OPTIMUS is retrained on the entire dataset to adjust for the new input size.

\begin{table}[ht]
\centering
\caption{Testing different context sizes}
{\footnotesize
\begin{tabular}{@{}l@{\hspace{6pt}}c@{\hspace{6pt}}c@{\hspace{6pt}}c@{\hspace{6pt}}c@{}}
\toprule
\textbf{Context size} & \textbf{F1 Score} & \textbf{AUROC} \\
\midrule
1  & 0.696 & 0.809  \\
2  & 0.745 & 0.850 \\
3 & \textbf{0.760} & \textbf{0.877}  \\
4  & 0.739  & 0.850 \\
5  & 0.689 & 0.817 \\
\bottomrule
\end{tabular}
}
\label{tab:ablation_context}
\end{table}

Table \ref{tab:ablation_context} presents the results of varying the context size. A context size of three was optimal, which is the size used for all other results in this paper. This is likely because three images provide a balance between robustness and variance in the temporal context given to the model. Performance decreases with context fewer than three due to reduced robustness to outliers. Conversely, performance drops with context more than three due to increased variability within the set, which complicates predictions.

\subsection{Change Measures}
For an ablation study, we evaluate using two different change measures, pivot score and Spearman coefficient. All of these were done on the Resnet-50 backbone, with Satlas weights, and with context size three.

\begin{table}[ht]
\centering
\caption{Testing different change measures}
{\footnotesize
\begin{tabular}{@{}l@{\hspace{6pt}}c@{\hspace{6pt}}c@{\hspace{6pt}}c@{\hspace{6pt}}c@{}}
\toprule
\textbf{Measure} &  \textbf{F1 Score} & \textbf{AUROC} \\
\midrule
Spearman & 0.748 & 0.860 \\
Pivot & \textbf{0.760} & \textbf{0.877} \\
\bottomrule
\end{tabular}
}
\label{tab:ablation_measure}
\end{table}

Pivot scores were slightly more effective than the Spearman coefficient in detecting progressive changes that aligned with human annotations. This may be because humans are better at identifying abrupt changes, which the pivot score captures more effectively.

\clearpage

\clearpage

\clearpage
\end{document}